\documentclass[journal]{IEEEtran}
\pdfoutput=1 
\usepackage{cite}
\usepackage{amsmath,amssymb,amsfonts}
\usepackage{algorithmic}
\usepackage{graphicx}
\usepackage{textcomp}

\usepackage{framed,multirow}
 \usepackage{url}
\usepackage{amssymb}
\usepackage{latexsym}
\usepackage{multirow}
\usepackage[table,xcdraw]{xcolor}
\usepackage{amsfonts}
\usepackage{url}

\usepackage{hyperref}
\definecolor{darkblue}{rgb}{0,0.08,0.45}
\definecolor{mblue}{rgb}{0.075,0.541,0.855}
\definecolor{nblue}{rgb}{0,0.263,0.576}
\definecolor{mydarkred}{rgb}{0.545, 0, 0}
\definecolor{warmblack}{rgb}{0.0, 0.26, 0.26}

\definecolor{mygray}{rgb}{0.93725,  0.93725,  0.93725}
\definecolor{aliceblue}{rgb}{0.94, 0.97, 1.0}
\definecolor{yaleblue}{rgb}{0.06, 0.3, 0.57}
\definecolor{ultramarine}{rgb}{0.07, 0.04, 0.56}
\definecolor{prussianblue}{rgb}{0.0, 0.19, 0.33}
\definecolor{oxfordblue}{rgb}{0.0, 0.13, 0.28}
\definecolor{coolblack}{rgb}{0.0, 0.18, 0.39}
\definecolor{darkpastelgreen}{rgb}{0.01, 0.75, 0.24}

\definecolor{britishracinggreen}{rgb}{0.0, 0.26, 0.15}
\definecolor{ceruleanblue}{rgb}{0.16, 0.32, 0.75}
\definecolor{darkcandyapplered}{rgb}{0.64, 0.0, 0.0}
\definecolor{forestgreen}{rgb}{0.0, 0.27, 0.13}

\usepackage{pifont}

\usepackage{amsmath}
\DeclareMathOperator{\sign}{sign}
\DeclareMathOperator{\argmin}{argmin}
\DeclareMathOperator{\argmax}{argmax}
\newcommand{\uargmin}[1]{\underset{#1}{\argmin}\;}

\usepackage[framemethod=TikZ]{mdframed}

\usepackage{tikz}

\usepackage[most]{tcolorbox}
\newtcolorbox{empheqboxed}{colback=aliceblue,
 colframe=white,
 width=.49\textwidth,
 sharpish corners,
 top=-2mm, 
 bottom=0pt
}

\usepackage{empheq}

\hypersetup{
  colorlinks   = true, 
  linkcolor    = blue, 
  citecolor   = blue 
}

\makeatletter
\def\thickhline{%
  \noalign{\ifnum0=`}\fi\hrule \@height \thickarrayrulewidth \futurelet
   \reserved@a\@xthickhline}
\def\@xthickhline{\ifx\reserved@a\thickhline
               \vskip\doublerulesep
               \vskip-\thickarrayrulewidth
             \fi
      \ifnum0=`{\fi}}
\makeatother

\newlength{\thickarrayrulewidth}
\ifCLASSINFOpdf
\else
\fi
\usepackage{framed}
\definecolor{shadecolor}{rgb}{0.9,0.9,0.9}

\begin{document}
%
\title{GraphXCOVID: Explainable Deep Graph \\ Diffusion Pseudo-Labelling  for Identifying \\ COVID-19  on Chest X-rays}
%
%
%
\author{Angelica I. Aviles-Rivero, 
\thanks{AI Aviles-Rivero is with the Department of Pure Mathematics and Mathematical Statistics, University of Cambridge, UK.  e-mail:
ai323@cam.ac.uk.}
Philip Sellars, \thanks{P. Sellars and C.-B. Sch\"{o}nlieb are with Department of Applied Mathematics and Theoretical Physics, University of Cambridge, UK. e-mail: \{ps644,cbs31\}@cam.ac.uk}
Carola-Bibiane Sch\"{o}nlieb
and Nicolas Papadakis  \thanks{N. Papadakis is with the Université de Bordeaux, IMB, Bordeaux INP, CNRS, UMR 5251. F 33-400, Talence, France. email: nicolas.papadakis@math.u-bordeaux.fr}
}

\markboth{}%
{}
%



\maketitle

\begin{abstract}
Can one learn to diagnosis COVID-19 under extreme minimal supervision?
Since the outbreak of the novel COVID-19 there has been a rush for developing Artificial Intelligence techniques for expert-level disease identification on Chest X-ray data.  In particular, the use of deep supervised learning has become the go-to paradigm. However, the performance of such models is heavily dependent on the availability of a large and representative labelled dataset. The creation of which is a heavily expensive and time consuming task, and especially imposes a great challenge for a novel disease. Semi-supervised learning has shown the ability to match the incredible performance of supervised models whilst requiring a small fraction of the labelled examples. This makes the semi supervised paradigm an attractive option for identifying COVID-19. In this work, we introduce a graph based deep semi-supervised framework for classifying COVID-19 from chest X-rays. Our framework introduces an optimisation model for graph diffusion that reinforces the natural relation among the tiny labelled set and the vast unlabelled data. We then connect the diffusion prediction output as pseudo-labels that are used in an iterative scheme in a deep net. We demonstrate, through our experiments, that our model is able to outperform the current leading supervised model with a tiny fraction of the labelled examples. Finally, we provide attention  maps to accommodate the radiologist's  mental model, better fitting their perceptual and cognitive abilities. These visualisation aims to assist the radiologist in  judging whether the diagnostic is correct or not, and in consequence to accelerate the decision.
\end{abstract}



%
\IEEEpeerreviewmaketitle

\section{Introduction} \label{sec:introduction}
\IEEEPARstart{S}{ince} the outbreak of the novel coranavirus disease 2019 (COVID-19), which is caused by severe acute respiratory syndrome coronavirus 2 (SARS-CoV-2), there have been more than 30 million confirmed infected cases and more than 1 million  deaths has been reported worldwide (as on September 1st). This threat has encouraged joint efforts to obtain accurate early detection of COVID-19 to try and limit the spread of the pandemic.

Whilst real-time reverse transcription polymerase chain reaction (RT-PCR) COVID-19 test is the current gold standard for diagnostis, this type of test has demonstrated several limits and burdens. Firstly, it is prone to false negatives that heavily rely on the sample acquisition characteristics including insufficient quantities and location (nasal, throat or sputum)~\cite{xie2020chest,wikramaratna2020estimating}. Secondly, the limitation of several world regions in obtaining fast accessibility to the test.
The use of imaging techniques, including computerised tomography and chest x-rays, has been suggested as a parallel option to RT-PCR test. {Clinical manifestation of COVID-19 is that of a respiratory infection, which is associated with viral pneumonia. Distinguishing viral pneumonia  from bacterial pneumonia is a challenging  task.
This task becomes even harder when a large number of suspected patients need to be screened. This time consuming task strains already limited medical resources, thus reducing the efficiency of the diagnostic.}

Computerised tomography (CT) has been a focus of attention in the literature for COVID-19  e.g.~\cite{zhou2020ct,chung2020ct}. However, the burden imposed in terms of infection control using CT scan suites and the inefficiencies relating to room decontamination and access restriction from several world regions make CT challenging to be used as a routinely basis -- despite its high sensitivity~\cite{fang2020sensitivity}.
Due to its wide availability and inexpensive screening, a great focus has been placed on Chest X-rays (CXR) for both clinical and AI areas e.g.~\cite{jacobi2020portable,wong2020frequency, cohen2020covid,covidnet}. These advantages make CXR a perfect alternative and complement to the RT-PCR. Despite the CXRs advantages, accurate interpretation still remains a challenge  ~\cite{folio2012chest}. This is because the accuracy of the interpretation relies on the radiologist's expertise level and there is still a substantial clinical error on the outcome~\cite{bruno2015understanding}. Therefore, there is an urge for fast automated evaluations of CXRs, to quickly explore the vast amount of  data which will save time in evaluating the diseases.

For the task of classifying COVID-19 using CXRs data, there has been a fast development of deep learning  techniques e.g.~\cite{narin2020automatic, apostolopoulos2020extracting, farooq2020covid, hemdan2020covidx, afshar2020covid}, in which supervised learning is the go-to paradigm. However, the performance of these techniques strongly rely on a large  and representative corpus of labelled data. In the medical domain and in particular with a new disease, this might be a strong assumption in the design of a solution, as it involves scarce annotations that  might contain strong human bias. The current leading supervised model for COVID, COVID-Net~\cite{covidnet}, has reported promising results with a sensitivity of 91\% for COVID-19. Hence there is still plenty of room for improvements, namely on how to use the vast amount of available unlabelled data to prevent labelling errors and uncertainties from affecting the classification output.

Motivated by the above limitations in current techniques, we address the following problem -- \textit{Can one get a robust classifier with performance higher or comparable to the current leading supervised technique for COVID-19  using far less labels?} To answer to this question we propose a deep semi-supervised framework to go beyond human bias and the limited amount of labelled data. {We remark that many deep SSL techniques  e.g.\cite{laine2016temporal,mixmatch,tarvainen2017mean,verma2019interpolation}, have only been considered for natural images. No work has evaluated the performance outside this domain, by considering the fundamental differences between natural and medical images~\cite{raghu2019transfusion}. This paper extends our work in~\cite{aviles2019graphx} with noticeable differences.} Firstly, we construct the graph based on the initial embeddings coming from a deep net, making for an accurate first construction. Secondly, we use our optimisation diffusion model as means for generating pseudo-labels that can be updated iteratively in a trained deep net. Moreover,  unlike the pseudo-label perspective of  \cite{lee2013pseudo}, our technique is a graph based approach and the pseudo-labels are computed by our diffusion model rather than the network. Our contributions are as follows:

\begin{itemize}
\setlength\itemsep{0em}
	\item We propose a deep semi-supervised framework, in which we highlight:
	\begin{itemize}
  \item An optimisation model with strong class priors for multi-class graph diffusion, which is based on normalised and non-smooth $p=1$ Dirichlet energy. Our method offers theoretical guarantees and efficient solving.
          \item  The connection of our diffusion model to the generation of meaningful pseudo-labels, which  avoids the current SSL trend on the use of consistency regularisation.
	\end{itemize}
We show that our framework reinforces the natural relation among the tiny labelled set and the vast unlabelled data. 
\noindent
	\item We evaluate our technique with several numerical, statistical and visual results using an unified dataset that contains highly diverse samples from different sources. To the best of our knowledge, this is the first graph based deep semi-supervised technique proposed for identifying COVID-19. Moreover, we also report explainable results from our prediction scores to assist and accelerate the radiologist diagnosis.
	\item We demonstrate that our technique reports higher sensitivity in COVID-19 and global performance than the current leading deep supervised technique for such application whilst requiring far less labelled data.
\end{itemize}

\section{Related Work}
The recent problem of classifying CXRs for COVID-19  has observed a fast growing in the literature. 
Existing techniques are reviewed in  this section.

\begin{figure*}[!t]
\centering
\includegraphics[width=1\textwidth]{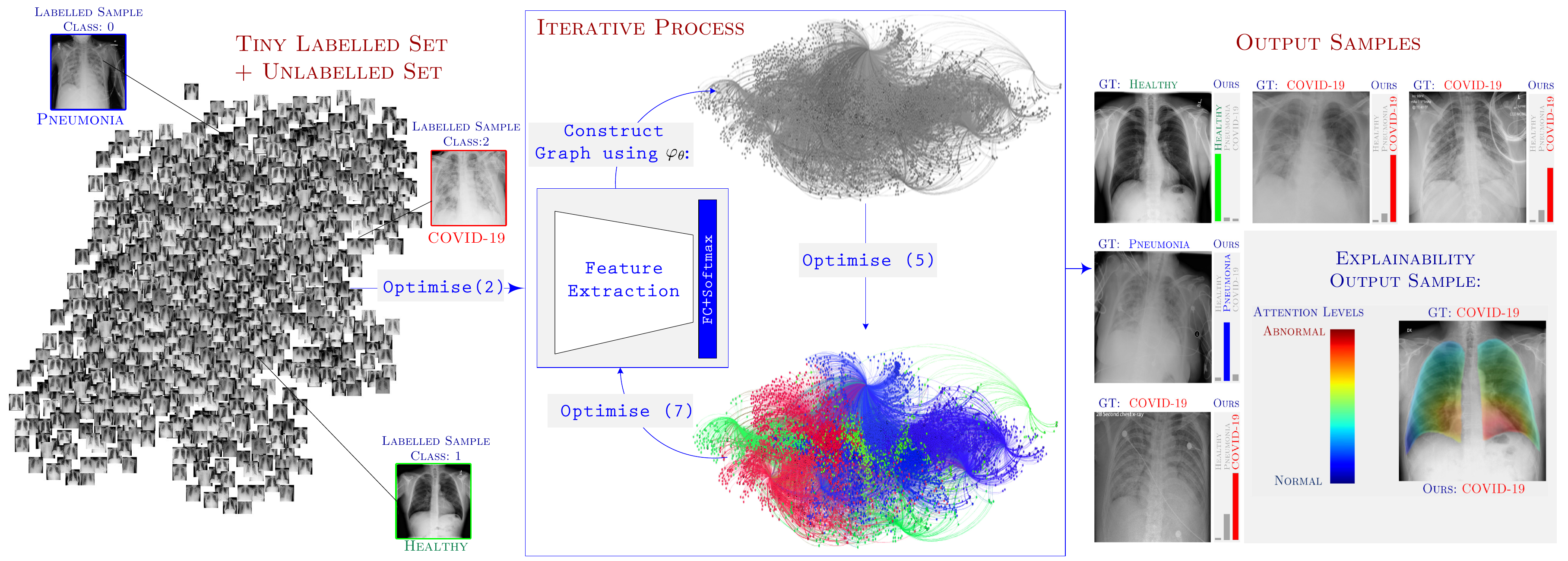}
\caption{Workflow of our proposed technique. We use a tiny labelled set and large unlabelled set. First our technique optimise over the labelled set using \eqref{LS_loss}, this seek to extract meaningful features to construct a strong graph. We then use our proposed diffusion model \eqref{pde}, which results in generated pseudo-labels than then are iteratively optimised using \eqref{Loss_totalUpdates}. The output is the confidence score per class and the attention map.}
\label{fig::teasearb}
\end{figure*}

\subsection{Chest X-ray Classification for COVID-19}
{The task of classifying CXRs has been widely investigated in the community. The go-to paradigm to address this problem is deep learning \cite{wang2017chestx,yao2018weakly, yao2018weakly,baltruschat2019}. The fast development of these techniques has been motivated by the release of several benchmarking datasets including ChestXray14\cite{wang2017chestx} and CheXpert~\cite{irvin2019chexpert}, in which a large number of annotated samples from different pathologies (classes) are contained in each dataset.  With new diseases such as COVID-19, where the annotated data is limited, one needs to rethink the whole design of the techniques. In this work, we motivate the power of semi-supervised learning for COVID-19. In what follows, we review  related COVID-19 and SSL techniques.}

The bulk of literature addressing {the task of classifying CXRs for} COVID-19 is largely based on deep supervised learning e.g.\cite{covidnet,narin2020automatic,apostolopoulos2020extracting,farooq2020covid,hemdan2020covidx,afshar2020covid,zhang2020covid}, {and  several techniques arise  every single day.}
Most of approaches apply  pre-trained off-the-shelf networks; in which  diverse generic architectures, including  ResNet~\cite{he2016deep}, DenseNet~\cite{densenet2016} and VGG~\cite{Simonyan15}.
Existing works thus leverage on fine tuned networks and networks trained from scratch on CXRs data. However, there are fundamental differences between natural and medical image classification including features and data size. Hence, as shown in~\cite{raghu2019transfusion}, transfer learning might offer little benefit to performance due to the over-parameterisation of standard models. {Moreover, the samples available for COVID-19 are scarse in comparison to other type of pneumonia, and one needs to deal with  highly imbalanced datasets.}

The key assumption of supervised techniques is a well-representative labelled dataset, and in new diseases, such as COVID-19, this core assumption is a strong one. Moreover,  the available annotations might be far from being a definite expression of ground truth \cite{kohli2017medical}.
Therefore, by using deep supervised learning techniques are prone to labelling error and uncertainty that adversely affect
the classification output. Although transfer learning~\cite{bar2015chest} and Generative Adversarial Networks~\cite{moradi2015machine} mitigate, at some level, the lack of a large and representative dataset, they weakly account for the mismatch between expert annotation and ground truth annotation, which is generated by human bias and uncertainty, and the performance can be limited due to the differences among datasets as demonstrated in~\cite{raghu2019transfusion}.

\color{black}
Motivated by the above-mentioned drawbacks in deep supervised learning and with the goal of generating a high sensitivity technique that largely decrease the need for large annotation set, \textit{we introduce a deep semi-supervised technique for the application to COVID-19 identification, which to our knowledge it is the first in its type.} In the next subsection we discuss  recent techniques in deep SSL for other domains and how they differ to ours. \color{black}

\subsection{Semi-Supervised Classification for Medical Images} \color{black}
Semi-supervised learning has been applied in the medical domain since its early developments, {in which model based techniques have been the main focus of attention  e.g.\cite{filipovych2011semi, chen2013inferring, dodero2014group, an2016semi, sun2016computerized}. These approaches have demonstrated the potential of SSL- but the generalisation of the feature space along with the computational requirements have raised  limitations. Recently, with the advent of deep learning, deep semi-supervised learning has been investigated. }

In the past few years, there has been a silent revolution in deep SSL techniques, e.g.~\cite{mixmatch,tarvainen2017mean,verma2019interpolation}, that have sought to combine the theoretical underpinning of SSL \cite{chapelle2009semi} with the generalisation and feature extraction of deep neural networks.
\color{black}
The largest trend for new deep SSL methods involves  \textit{consistency regularisation} \cite{laine2016temporal,mixmatch,tarvainen2017mean,verma2019interpolation}. The core idea of this perspective is to, for both labelled and unlabelled examples, induce perturbations $\delta$ and then add a regularisation term to the loss function such that the prediction of the model is invariant to the perturbation, i.e.  $f(x+\delta) = f(x)$. The main variety between approaches stem from how to generate  perturbations $\delta$.
The definition of perturbations is indeed a complex problem, and no work has ever evaluated the performance of consistency based approaches outside the natural image domain.

To the best of our knowledge, there has been no deep SSL approach proposed for COVID-19 analysis. The potential performance of  SSL has nevertheless been shown in our prior work \cite{aviles2019graphx} on  CXRs analysis. We readily competed with SOTA supervised techniques for identification of several pathologies in CXRs, using a small fraction of the available labels. In this work, we extend this framework  
to build upon the concept of pseudo-labels, first introduced in \cite{lee2013pseudo}. However, compared to that initial work, there are now several key differences since a graph based approach is considered and  the pseudo-labels are predicted from our diffusion model rather than the deep net.

\color{black}
\section{GraphX-COVID Framework}
This section presents the three parts of our proposed technique: i) data representation and robust graph construction, ii) optimisation model for graph diffusion and iii)  driving optimisation that connects our diffusion model with deep nets. The overview of our GraphXCOVID is illustrated in Fig. \ref{fig::teasearb}.

\medskip
\noindent
\textcolor{black}{\textbf{Problem Definition.}} Given a small amount of labelled data $D_L =\{ (x_h ,y_h) \}_{h=1}^{l}$ with provided labels $\mathcal{L} = \{1,..,L\}$  and $y_h \in \mathcal{L}$, and a large amount of unlabelled data  $X_u = \{ x_k \}_{m=l+1}^{n}$. The whole set of data is thus $X= X_L \cup X_U$, where $X_L= \{x_1,...,x_l\}$. We seek to infer a function $f: \mathcal{X}^{n} \mapsto \mathcal{Y}^{n}$ such that $f$ gets a good estimate for  $\{x_k \}_{m=l+1}^{n}$
with minimum generalisation error.

\medskip
In particular, in a deep semi-supervised setting, one seeks to minimise a functional of the form:
\begin{equation} \label{generalSSL}
  \min_{\theta}\textcolor{darkblue}{\overbrace{\sum_{(x,y)\in D_L }\mathcal{L_S}(x,y;\theta)}^{\text{labelled set}}} + \gamma \textcolor{mydarkred}{\overbrace{\sum_{x\in X_u } \mathcal{L_U}(x;\theta)}^{\text{unlabelled set}}},
\end{equation}

\noindent
where \textcolor{darkblue}{$\mathcal{L_S}$} is the per example loss for the labelled set (e.g. standard cross-entropy) and \textcolor{mydarkred}{$\mathcal{L_U}$} denotes a loss defined on the unlabelled set (e.g. consistency loss). Moreover, $\gamma \in \mathbb{R}^+$ is a weighting parameter to balance the two terms, and $\theta$ is the network parameters to estimate.

The current go-to perspective in deep SSL is to apply consistency regularisation  for \textcolor{mydarkred}{$\mathcal{L_U}$} in~\eqref{generalSSL}, which enforces invariant network predictions with respect to perturbations on the unlabelled data $X_u$, e.g. \cite{laine2016temporal,mixmatch,tarvainen2017mean,verma2019interpolation}
However, the definition of such  $\delta$-pertubations, e.g.  flip-and-shift, rotate, posterise and sharpness,  is not trivial. 
In this work, we avoid the explicit definition of such $\delta$ by taking a \textit{proxy-based perspective}. In particular, we rely on the concept of pseudo-labels~\cite{lee2013pseudo} $\hat y_i$ for images of the unlabelled set $x_i\in X_u$. In this work, the pseudo-labels are generated by optimising our graph-based model.
Our framework  is an iterative two-part technique, the first part concerns  pseudo-label generation, including graph representation (see Subsection A)   and label diffusion (Subsection B). The second step deals with the update of the generated pseudo-labels (Subsection C).

\subsection{Feature Extraction \& Graph Construction}

The most common data representation is a Euclidean or grid-like structure. We rather define our data in a non-Euclidean domain with a graph.
This framework offers different benefits including mathematical properties such as sparseness which allows for fast computation, and the ability to correct  initially mislabelled data by smoothing the embeddings. We represent the dataset $X$ as a graph, where each node is an image, to produce pseudo-labels.  Then, unlike pure model-based approaches or pure deep learning techniques, we introduce a hybrid model -- that is, a combination between a model-based (energy model) and a deep learning framework.

A deep network $f_\theta$  is considered for updating the pseudo-labels generated by our optimisation model.
It is initialised from the tiny labelled set $(x,y) \sim  D_L$ by  minimising:
\begin{empheqboxed}
\begin{align} \label{LS_loss}
{
  \mathcal{L_S}(X_L,Y_L;\theta) := \sum_{h=1}^{l} \ell_H(f_{\theta}(x_h),y_h),}
\end{align}
\end{empheqboxed}

\noindent
where the loss function $\ell_H$ is cross entropy, which is the most common choice for classification tasks. This optimisation process  only involves the provided small labelled set to construct the initial graph (i.e. it is run once as initialisation).

\begin{figure}[!t]
\centering
\includegraphics[width=0.45\textwidth]{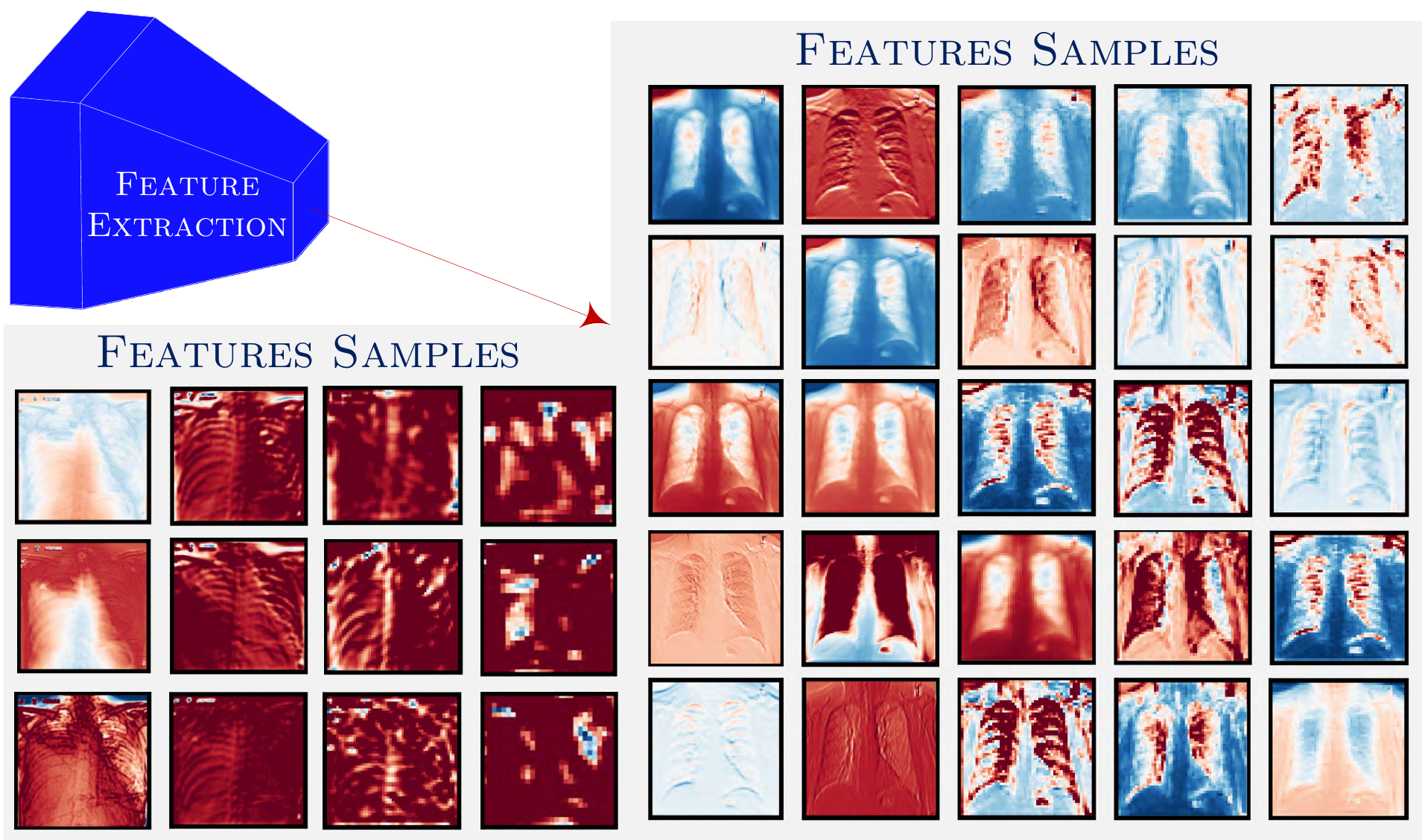}
\caption{
Visualisation of feature  extracted from Chest X-ray data.\vspace{-0.2cm}}
\label{fig::teasear}
\end{figure}

More precisely, a given set of data (or features) can be represented as an undirected weighted graph by $\mathcal{G}=(\mathcal{V},\mathcal{E},\mathcal{W})$ composed of  $n$ nodes $\mathcal{V} = \{ v_1, ... , v_n\}$, which are connected by edges $\mathcal{E}= \{\{v_i,v_j\} : v_i,v_j\in\mathcal{V}\}$ 
with weights $w_{ij}=S(i,j)\geq 0$ that correspond to some similarity measure $S$ between the features of nodes $i \in \mathcal{V}$ and $j \in \mathcal{V}$, and $w_{ij}=0$  if  $(i,j)  \notin  \mathcal{E}$. In our setting, a node $v_i$ represents an image of the set $x_i\in X$. A fundamental question when using graph based approaches is how to effectively extract representative features to construct a robust graph. In order to avoid the sensitive task of hand-crafted feature selection, we  rely on  embeddings automatically produced by \eqref{LS_loss}.  Hence we reduce the potentially large generalisation error that appear when extracting hand-crafted features on a small training set $X_L$.

We consider as feature extractor the function $\varphi_\theta : \mathcal{X}\rightarrow \mathbb{R}^P$, given by the bottleneck of the current network $f_\theta$, that maps the input to some feature space of dimension $P$.
To construct our graph, we compute, for each sample, the set of descriptors by  $\mathbf{c_i}=\varphi_\theta(x_i)\in\mathbb{R}^P$ for $x_i \in X$, and connections are created through the k nearest  neighbours (k-NN) approach in $\mathbb{R}^P$. An illustration of extracted features is given in Fig. \ref{fig::teasear}.

Notice that the model $f_\theta$ and the feature extractor $\varphi_\theta$ are updated (subsection C) through the exposition to pseudo-labels of  unlabelled data. 
%
The graph thus evolves along our process. We now detail how pseudo-labels are obtained on the graph by following a transductive strategy.

\subsection{Label Diffusion as Pseudo-Labelling}
Pseudo-labels are estimated through a diffusion process on the whole graph containing both labeled and unlabelled examples. For each image, our method also assigns a score reflecting the uncertainty of the produced pseudo-label (see subsection C).
To do this, we first generate the pseudo-labels through an optimisation model based on  the normalised graph p-Dirichlet energy expressed as:

\begin{equation}\label{model}
\Delta_p (u)=\sum_{i,j} w_{ij}\left\| \frac{u_i}{d_i^{1/p}}-\frac{u_j}{d_j^{1/p}}\right\|^p, \vspace{0cm} 
p\geq 1,\,d_i=\sum_j w_{ij}>0, 
\end{equation}

\noindent
where the degree of the node $i$ is denoted by $d_i$, and the weights are computed from the descriptors described in previous section.
The minimisation of this energy allows the diffusion of a labelling variable  $u$.
Whilst techniques in this line have been reported in the medical domain  e.g.\cite{chen2013inferring,dodero2014group,wang2016progressive} and in the pure machine learning  community e.g.~\cite{belkin2003Laplacian,zhou2004learning}, they only seek to use eigenfunctions of a normalised Dirichlet energy based on the graph Laplacian for $p=2$   or only approximate $p\to 1$. However, latter machine learning works  \cite{Buhler2009} has demonstrated that using the non smooth $p=1$ Dirichlet energy (related to total variation) achieves better performance for label propagation.

With this motivation in mind, we introduce an optimisation model based on the normalised and non smooth  $p=1$ Dirichlet energy. In the case $p=1$, model \eqref{model} is thus defined as $\Delta_1(u)=|WD^{-1}u|$, where  $D$ is the  diagonal matrix containing the degrees $d_i$ and the $m\times n$ matrix  $W$ encodes the $m$ edges in the graph. Each of these edges is represented on a different line $i$ of the matrix $W$,  with the value $w_{ij}$ (resp. $-w_{ij}$) on the column $i$ (resp. $j$).

We now detail our multi-class model that will be applied to $L=3$ classes: \textit{0: Healthy}, \textit{1: Pneumonia} and \textit{2: COVID-19}. For each class $k=1\cdots L$, we set a variable $u^k$ that contains the node values for class $k
$ and denote  ${\bf u}= [u^1,\cdots u^L]$.
For unlabelled nodes $i>l$, we  couple the $L$ variables with the  constraint: $\sum_{k=1}^L u^k_i=0,\, \forall i>l$.  We make the standard assumption that there exists a non empty set of labelled nodes $\mathcal I_k\subset\{1\cdots l\}$ for each class $k$. For these nodes, we set $u^k_i\geq \epsilon$ if $i\in\mathcal I_k$ (positive response for the class), and $u^{k'}_i\leq -\epsilon$ if  $i\in\mathcal I_k$ and $k'\neq k$ (negative output for the other classes).

Under such constraints, we seek to minimise the  multi-class functional \cite{bresson2013multiclass} that contains the sum of normalised ratios :
\begin{equation}\label{RQ2}\min_{||{\bf u}||=1}\sum_{k=1}^L\frac{\Delta_1  u^k}{|u^k|}.\end{equation}
We consider an iterative scheme  to optimise this problem:

{\small
\begin{empheqboxed}
\begin{align}\label{pde}
\begin{split}
{\bf u}^{(t+1)}&
=\uargmin{{\bf u}} \frac{\| {\bf u}-{\bf u}^{(t)}\|^2}{2\Delta t }\\&+\sum_{k=1}^L\left( \Delta_1(u^k)-\frac{\Delta_1 (u^{k,(t)})}{|u^{k,(t)}|}\langle \sign(u^{k,(t)}), u^k\rangle \right),
\end{split}
\end{align}
\end{empheqboxed}
}

\noindent
where $t$ is a time index associated to the step $\Delta t>0$.
This process diffuses information from labelled nodes to unlabelled ones.
To avoid trivial solutions~\cite{hein2013total,Feld2019}, we apply  shifting ${u}^{k,(t+1)}={u}^{k,(t+1)}-\textrm{median}({u}^{k,(t+1)})$ and  normalisation  ${\bf u}^{(t+1)}={\bf u}^{(t+1)}/||{\bf u}^{(t+1)}||$ steps at the end of each iteration.

From \eqref{pde}, one can see that  the solution $u^{k,(t+1)}$ satisfies:

\begin{equation}\label{decrease}
\begin{split}
\sum_{k=1}^L\Delta_1 (u^{k,(t+1)})&\leq \sum_{k=1}^L\frac{\Delta_1 (u^{k,(t)})}{|u^{k,(t)}|}\langle \sign(u^{k,(t)}), u^{k,(t+1)}\rangle\\&\leq\sum_{k=1}^L \frac{\Delta_1 (u^{k,(t)})}{|u^{k,(t)}|}|u^{k,(t+1)}|,
\end{split}
\end{equation}
\noindent
so that we get a reduction of the normalised ratio $\Delta_1 (u^{k,(t)})/|u^{k,(t)}|$ along iterations $k$. When $L=1$, the scheme $u^{k,(t)}$ converges to {a local minima of \eqref{RQ2} which corresponds to} a bivalued function that naturally segments the graph~\cite{Feld2019}.
{In the general case $L>1$, the convergence to a local minima of \eqref{RQ2} can be ensured by using a modification of the scheme \eqref{pde} as proposed in \cite{rangapuram2014tight}. However, such adaptation comes at the cost of an important additional computational cost. Even of there is no theoretical guarantee for that, we observe a monotonous decrease of \eqref{RQ2} with the scheme \eqref{pde}. As a consequence, we suggest to use the flow \eqref{pde} that presents an acceptable trade-off between theoretical and practical aspects.}

Once ${\bf u}
^k$ has converged to some ${\bf u}
^*=[u ^{*,1},\cdots, u^{*,L}]$, {the label of} each node is finally given by $\widehat{y}_i=\argmax_{j} u_i ^{*,j}$. In practice, our model \eqref{pde} is solved using an accelerated primal dual algorithm  \cite{chambolle2011first}.
Our generated pseudo-labels are denoted $\widehat{Y}_U = \{\widehat{y}_k \}_{k=l+1}^{n}$ and used to update the classification network $f_\theta$ as explained in the next section.

{We remark that there are several differences between our pseudo-label approach and that of \cite{lee2013pseudo,aviles2019graphx}. Firstly and unlike~\cite{aviles2019graphx}, our work generates and updates the embeddings from a deep net to construct, since the beginning, a stronger  graph. Secondly, our optimisation model generates pseudo-labels, outside the deep net, and then they are iteratively updated from a deep net. Thirdly, our current model is designed to generate highly certain pseudo-labels since early stages by integrating an uncertainty measure, and also is equipped with a class balance term (see Section C). With respect to~\cite{lee2013pseudo}, our model follows different principle when generating the pseudo-labels. Firstly, unlike \cite{lee2013pseudo}, our technique
is a graph based model. Secondly, ~\cite{lee2013pseudo} generates
the pseudo-labels directly from the deep net (i.e. inside the network) by taking maximum predicted
probability from the network. On the other hand, we generate the pseudo-labels from our new optimisation model (drawn from
the normalised graph p-Dirichlet energy) which is decoupled from
the network (i.e. outside the network).   }

\subsection{Deep Graph Pseudo-Labelling Update} 
Although our graph diffusion model \eqref{pde} generates relevant pseudo-labels \cite{aviles2019graphx}, one can further decrease the uncertainty over time. This can be achieved if we consider the two major bottlenecks in real-world problems. A first prevalent issue in the medical domain is to face highly imbalanced class samples. As illustrated in Fig. \ref{fig::dataset}, this is particularly true in the COVIDx dataset. The second one is to deal with 
inferred pseudo-labels with different levels of  uncertainty. As it has been shown in several works e.g.~\cite{he2013imbalanced,shi2018transductive,iscen2019label,sellars2020two}, these problems can be mitigated by weighting the importance over the inferred pseudo-labels and the classes.

The problem of classifying with a highly imbalanced dataset has been widely studied in the literature, e.g.~\cite{kukar1998cost,he2013imbalanced}.
We apply a common strategy for imbalanced class population~\cite{he2013imbalanced,bookImbalanceHerrera} and add a weighting factor  inversely proportional to the effective number of samples for class $k: \omega_k \propto 1/E_n, n\in \mathbb{Z}_{>0}$ where $E_n$ is the total number of samples. For the second problem,  we associate an uncertainty weighing factor, $\upsilon_i$, to each $u_{\hat{y}_i}$ generated in the diffusion process. We use entropy as the measure for uncertainty, given by  $\upsilon_i = 1 - (H(u_{\hat{y}_i})/\log (L)) $, where $H$ refers to the entropy and  $u_{\hat{y}_i}$ is normalised beforehand with respect to the values in $u_i^*$.

We finally define the main driving optimisation (i.e. estimation of network parameters) as:

\begin{empheqboxed}
\begin{equation}\label{Loss_totalUpdates}
{
  \min_{\theta}\sum_{i=1}^{l} \omega_{y_i} \ell_H(f_{\theta}(x_i),y_i) +\sum_{i=l+1}^{n} \upsilon_i\omega_{\widehat{y}_i} \ell_H(f_{\theta}(x_i),\widehat{y}_i)},
\end{equation}
\end{empheqboxed}
\noindent
the loss in \eqref{Loss_totalUpdates} is connected with model \eqref{generalSSL} but unlike the typical consistency loss, for $\mathcal{L_U}$, we are using the philosophy of pseudo-labels which are generated by our diffusion model.

Let us summarise the overall process. We first optimise \eqref{LS_loss} for a set of epochs, this serves for extracting the embeddings from the deep net to construct a graph. We then perform \eqref{pde} to diffuse the small labelled set to the unlabelled data. The output of this process is the generation of pseudo-labels for the unlabelled set. These pseudo-labels are then used to optimise \eqref{Loss_totalUpdates}, which in turn updates the model parameters. The whole process (feature extraction, graph update, pseudo label diffusion, network update) is then iterated.

\section{Experimental Results} \color{black}
In this section, we detail the set of experiments conducted to validate our  technique.

\subsection{Dataset Description}
We evaluate our approach on the COVIDx Dataset which was introduced in \cite{covidnet}. The dataset is composed of a total of 15,254 CXR images. The official partition only considers 13,975 CXR images across 13,870 patients, with a training set composed of 13,675 images and a test test of 300 (1579 for the test set on the full dataset). This dataset is, up to our knowledge, the largest and most diverse one for COVID-19. COVIDx indeed merges five different datasets repositories:  COVID-19 image data collection~\cite{cohen2020covid},  Actualmed COVID-19 Chest X-ray Dataset Initiative  \cite{ActualMedDataset},  COVID-19 Chest X-ray Dataset Initiative \cite{Fig1Dataset}, RSNA Pneumonia Challenge dataset ~\cite{RSNA,wang2017chestx} and COVID-19 Radiography Database\cite{RadCovid5}.

\begin{figure}[!t]
\centering
\includegraphics[width=0.47\textwidth]{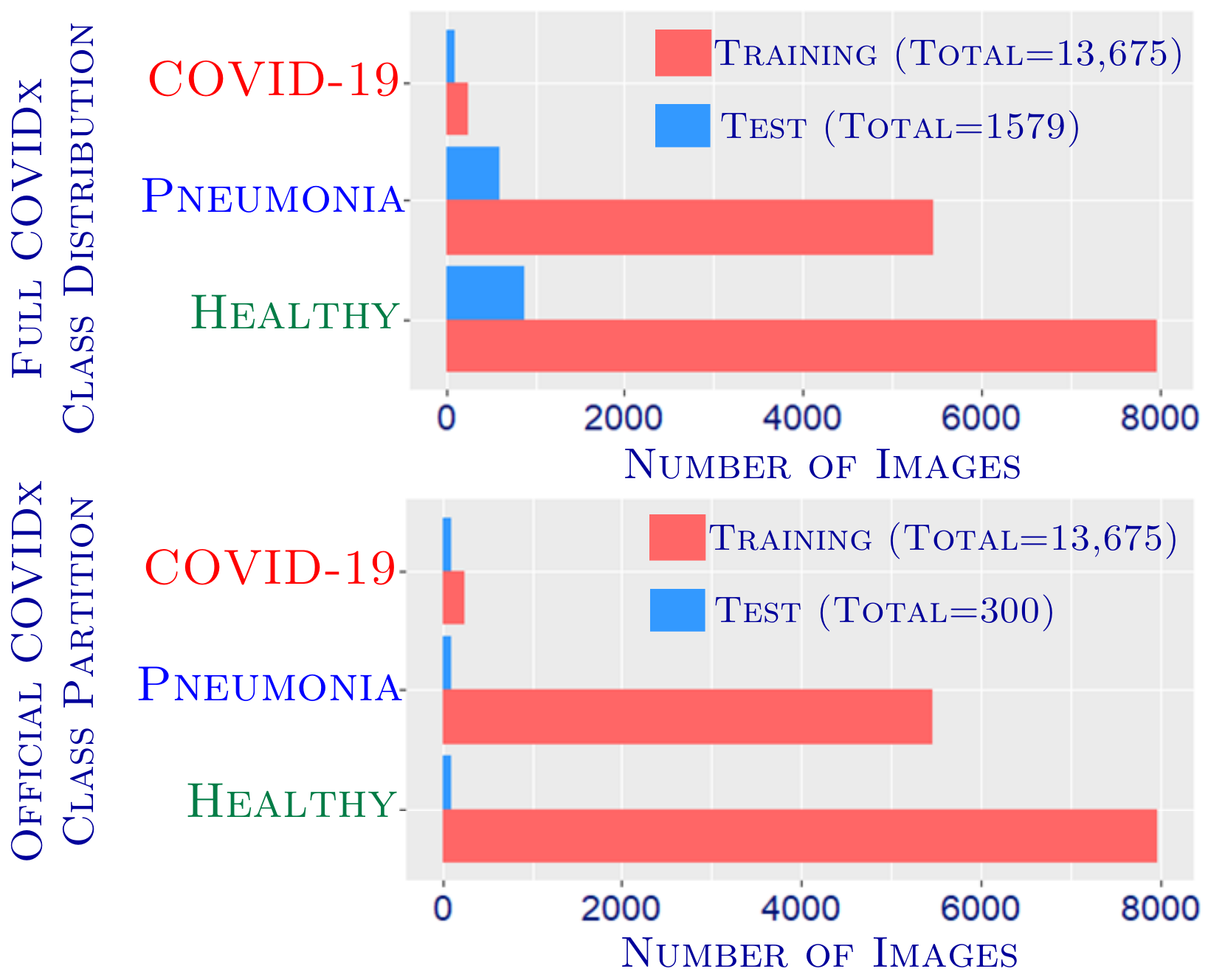}
\caption{Class distribution of COVIDx dataset. From top to bottom: samples/class distribution over the full COVIDx dataset and the official partition from COVIDx.   We conducted experiments in both sets.
This dataset imposes several challenges due to the highly imbalanced classes, in which COVID-19 samples are significantly less than the Pneumonia and Healthy classes. 
}
\label{fig::dataset}
\end{figure}

The COVIDx dataset contains three classes: Healthy, Pneumonia and COVID-19. The class breakdown, for the  full and official partition CXR images,  is illustrated in Fig. \ref{fig::dataset}. As it can be observed from these plots,  it is a highly imbalanced dataset, in which COVID-19 samples are much smaller than for the other other two classes.

{Moreover, to further support the performance of our technique, we use an external dataset to evaluate the generalisation to out-of-distribution samples. To do this, we use the external dataset BIMCV-COVID19~\cite{de2020bimcv} , which is composed of images from 11 hospitals from the Valencian Region, Spain. We randomly selected a subset of 200 patient-level samples covering all hospitals, where  75\% were COVID-19 confirmed cases as one is interested in show generalisation for the target disease.}


\begin{table*}[]
\caption{\label{table:comparisonOfficialPartition2}Numerical comparison of our technique vs fully supervised approaches. The results report per class metrics, including sensitivity, positive predictive value and F1-scores along with the overall accuracy. Our technique readily competes with all supervised techniques whilst using far less labelled data. $\dag$ denotes the score reported in~\cite{covidnet}.}
\centering
\resizebox{1\textwidth}{!}{
\begin{tabular}{ccclcccc}
\rowcolor[HTML]{EFEFEF}
\cellcolor[HTML]{EFEFEF} & \multicolumn{2}{c}{\cellcolor[HTML]{EFEFEF}\textsc{Learning Paradigm}} & \multicolumn{1}{c}{\cellcolor[HTML]{EFEFEF}} & \cellcolor[HTML]{EFEFEF} & \cellcolor[HTML]{EFEFEF} & \cellcolor[HTML]{EFEFEF} & \cellcolor[HTML]{EFEFEF} \\
\rowcolor[HTML]{EFEFEF}
\multirow{-2}{*}{\cellcolor[HTML]{EFEFEF}\textsc{Technique}} & \begin{tabular}[c]{@{}c@{}}\textsc{SL}\end{tabular} & \begin{tabular}[c]{@{}c@{}}\textsc{SSL}\end{tabular} & \multicolumn{1}{c}{\multirow{-2}{*}{\cellcolor[HTML]{EFEFEF}\textsc{Class}}} & \multirow{-2}{*}{\cellcolor[HTML]{EFEFEF}\begin{tabular}[c]{@{}c@{}}\textsc{Positive Predictive}\\ \textsc{Value (PPV)}\end{tabular}} &\multirow{-2}{*}{\cellcolor[HTML]{EFEFEF}\textsc{Sensitivity}} &  \multirow{-2}{*}{\cellcolor[HTML]{EFEFEF}\textsc{F1-Scores}} & \multirow{-2}{*}{\cellcolor[HTML]{EFEFEF}\begin{tabular}[c]{@{}c@{}}\textsc{Accuracy}\\ in $10^{-2}$\end{tabular}} \\ \hline
 &  &  & \multicolumn{1}{c}{\textcolor{forestgreen}{\textsc{Healthy}}} &  0.83 & 0.95 &  0.88   & \\
 &  &  & \textcolor{blue}{\textsc{Pneumonia}} & 0.76 &0.89 &  0.82 &  \\
\multirow{-3}{*}{\textsc{VGG-16}~\cite{Simonyan15}} & \multirow{-3}{*}{\checkmark} & \multirow{-3}{*}{} & \textcolor{red}{COVID-19} & \multirow{-3}{*}{}0.88 & \multirow{-3}{*}{}0.60 & \multirow{-3}{*}{}0.71 & \multirow{-3}{*}{81.3} \\ \hline
 &  &  & \multicolumn{1}{c}{\textcolor{forestgreen}{\textsc{Healthy}}} & 0.84 &0.91 &  0.87  &  \\
 &  &  & \textcolor{blue}{\textsc{Pneumonia}} & 0.85 &0.90 &  0.87  &  \\
\multirow{-3}{*}{\textsc{ResNet-18~\cite{he2016deep}}} & \multirow{-3}{*}{\checkmark} & \multirow{-3}{*}{} & \textcolor{red}{COVID-19} & \multirow{-3}{*}{}0.90 & \multirow{-3}{*}{}0.79 & \multirow{-3}{*}{}0.84 & \multirow{-3}{*}{86.7} \\ \hline
 &  &  & \multicolumn{1}{c}{\textcolor{forestgreen}{\textsc{Healthy}}} & 0.90 & 0.90 & 0.90  &  \\
 &  &  & \textcolor{blue}{\textsc{Pneumonia}}&  0.88 &  0.85  & 0.87 &  \\
\multirow{-3}{*}{\textsc{Pseudo-Labelling}~\cite{lee2013pseudo}} & \multirow{-3}{*}{} & \multirow{-3}{*}{\checkmark} & \textcolor{red}{COVID-19} & \multirow{-3}{*}{}0.84 & \multirow{-3}{*}{}0.86 & \multirow{-3}{*}{}0.84 & \multirow{-3}{*}{87.3} \\ \hline
 &  &  & \multicolumn{1}{c}{\textcolor{forestgreen}{\textsc{Healthy}}} & 0.88  &\cellcolor[HTML]{9AFF99} 0.97 &  0.92 &  \\
 &  &  & \textcolor{blue}{\textsc{Pneumonia}} & 0.87  & 0.91 &  0.89 &  90.0\\
\multirow{-3}{*}{\textsc{ResNet-50~\cite{he2016deep}}} & \multirow{-3}{*}{\checkmark} & \multirow{-3}{*}{} & \textcolor{red}{COVID-19} & \multirow{-3}{*}{}0.95 & \multirow{-3}{*}{}0.82 & \multirow{-3}{*}{}0.88  &  \multirow{-3}{*}{} 90.6$^\dag$\\ \hline
 &  &  & \multicolumn{1}{c}{\textcolor{forestgreen}{\textsc{Healthy}}} & 0.93 & 0.91 & 0.92  &  \\
 &  &  & \textcolor{blue}{\textsc{Pneumonia}}    & 0.90 & 0.92 & 0.91 &  \\
\multirow{-3}{*}{\textsc{InceptionV3}~\cite{szegedy2016rethinking}} & \multirow{-3}{*}{\checkmark} & \multirow{-3}{*}{} & \textcolor{red}{COVID-19} & \multirow{-3}{*}{}0.92 & \multirow{-3}{*}{}0.88 & \multirow{-3}{*}{}0.90 & \multirow{-3}{*}{91.0} \\ \hline
 &  &  & \multicolumn{1}{c}{\textcolor{forestgreen}{\textsc{Healthy}}} & 0.92 & 0.94 & 0.93   &  \\
 &  &  & \textcolor{blue}{\textsc{Pneumonia}} & 0.90   & 0.93 & 0.91 &  \\
\multirow{-3}{*}{\textsc{DenseNet-121}~\cite{densenet2016}} & \multirow{-3}{*}{\checkmark} & \multirow{-3}{*}{} & \textcolor{red}{COVID-19} & \multirow{-3}{*}{}0.93 & \multirow{-3}{*}{}0.88 & \multirow{-3}{*}{}0.90 & \multirow{-3}{*}{91.7} \\ \hline
 &  &  & \multicolumn{1}{c}{\textcolor{forestgreen}{\textsc{Healthy}}} & 0.90  & 0.95 &  0.93 &  \\
 &  &  & \textcolor{blue}{\textsc{Pneumonia}}& 0.91  &  0.94  & \cellcolor[HTML]{9AFF99}0.93 &  \\
\multirow{-3}{*}{\textsc{COVID-Net}~\cite{covidnet}} & \multirow{-3}{*}{\checkmark} & \multirow{-3}{*}{} & \textcolor{red}{COVID-19} & \multirow{-3}{*}{}\cellcolor[HTML]{9AFF99}0.98 & \multirow{-3}{*}{}0.91 & \multirow{-3}{*}{}\cellcolor[HTML]{9AFF99}0.95 & \multirow{-3}{*}{93.3} \\ \hline
 &  &  & \multicolumn{1}{c}{\textcolor{forestgreen}{\textsc{Healthy}}} & \cellcolor[HTML]{9AFF99}0.93 & 0.92  & \cellcolor[HTML]{9AFF99} 0.98 &  \\
 &  &  & \textcolor{blue}{\textsc{Pneumonia}} & \cellcolor[HTML]{9AFF99} 0.96 & \cellcolor[HTML]{9AFF99}0.96  & 0.92 &  \\
\multirow{-3}{*}{\textsc{GraphXCOVID}} & \multirow{-3}{*}{} & \multirow{-3}{*}{\checkmark} & \textcolor{red}{COVID-19} & \multirow{-3}{*}{}0.95 & \multirow{-3}{*}{}\cellcolor[HTML]{9AFF99}0.94 & \multirow{-3}{*}{}\cellcolor[HTML]{9AFF99}0.95 & \multirow{-3}{*}{\textbf{94.6}} \\ \hline
\end{tabular}}
\end{table*}

\subsection{Evaluation Methodology \& Implementation Details} \color{black}
We validate our proposed technique as follows. First we evaluate the performance of our approach compared to supervised techniques including the leading fully supervised paper in the field COVID-Net~\cite{covidnet}. These comparisons include VGG-16~\cite{Simonyan15},  ResNet-18 and ResNet-50~\cite{he2016deep}, InceptionV3~\cite{szegedy2016rethinking} and DenseNet-121~\cite{densenet2016}. The selection of these architectures follows the same line of motivation as in~\cite{covidnet}: they  offer a clear advantage for dealing with the unique traits of COVID. Our experiments are then conducted using: i) the official partition, in which the test set is 300 samples split evenly across the classes; ii) the full COVIDx dataset, in which the main differences with respect i) is that the test set is composed of 1,579 samples (see Fig.~\ref{fig::dataset}); and iii) an additional random partition.
We ran all the experiments under the same conditions, and followed standard pre-processing protocol to normalise the images to have zero mean and unit variance. The images we resized to the resolution 480$\times$480.

The evaluation is addressed  from  both  qualitative  and  quantitative  points  of  view. The former is based on visual outputs of our classification. The latter present the per class computation  of  sensitivity, positive predictive value and F1-scores.  The overall performance is computed in terms of accuracy and error rate.  Furthermore,  for sake of completeness and guided by the field of estimation statistics, we report along with the error rate the confidence intervals (95\%) of all techniques. Finally,   we performed a data ablation study of our SSL method by using $10\%, 20\% \text{ and } 30\%$ of labels.

We now give implementation details. For the  COVID-Net~\cite{covidnet} technique, we used the implementation and parameters provided by the authors. In particular, we considered the latest suggested model COVIDNet-CXR3-B. For the compared techniques, we used weight decay= 5e-4, momentum= 0.9 and learning rate 1e-2 (1e-3 for~\cite{lee2013pseudo}, and ResNet-18 for a fair comparison).
For our technique, the k-NN neighborhood graph has been built with $k=50$.  A ResNet-18 architecture has been used for the deep network $f_\theta$. In practice, we used a total number of epochs of 210 with and a weight decay of $2\times 10^{-4}$
and learning rate was set to 5e-2  decreasing with  cosine annealing.
Furthermore, we follow standard protocol in semi-supervised learning to report our results, we
randomly select the labelled samples over five repeated times, that is- one has five different splits. We then report the mean error over the splits.
All techniques were implemented in PyTorch and using Stochastic Gradient Descent (SGD) as optimiser.

\begin{table}[]
\centering
\caption{Performance comparison of COVID-Net and our technique using the full dataset.} 
\label{table:comparisonFullPartition}
\resizebox{0.47\textwidth}{!}{
\begin{tabular}{lccc}
\cline{2-4}
 & \cellcolor[HTML]{EFEFEF}\textcolor{forestgreen}{\textsc{Healthy}} & \multicolumn{1}{l}{\cellcolor[HTML]{EFEFEF}\textcolor{blue}{\textsc{Pneumonia}}} & \multicolumn{1}{l}{\cellcolor[HTML]{EFEFEF}\textcolor{red}{COVID-19}} \\ \cline{2-4}
\multirow{-2}{*}{}
 & \multicolumn{3}{c}{\cellcolor[HTML]{EFEFEF}\textsc{Positive Predictive Value} (PPV)} \\ \hline
COVID-Net & 0.96 & 0.91 & 0.93 \\
GraphXCOVID & \cellcolor[HTML]{9AFF99}0.97 & \cellcolor[HTML]{9AFF99}0.92 & \cellcolor[HTML]{9AFF99}0.95 \\ \hline
& \multicolumn{3}{c}{\cellcolor[HTML]{EFEFEF}\textsc{Sensitivity}} \\ \hline
COVID-Net & 0.93 & \cellcolor[HTML]{9AFF99}0.95 & 0.91 \\
GraphXCOVID & \cellcolor[HTML]{9AFF99}0.94 & \cellcolor[HTML]{9AFF99}0.95 & \cellcolor[HTML]{9AFF99}0.93 \\ \hline
 & \multicolumn{3}{c}{\cellcolor[HTML]{EFEFEF}\textsc{F1-Scores}} \\ \hline
COVID-Net & 0.95 & 0.93 & 0.92 \\
GraphXCOVID & \cellcolor[HTML]{9AFF99}0.96 & \cellcolor[HTML]{9AFF99}0.94 & \cellcolor[HTML]{9AFF99}0.94 \\ \hline
\end{tabular}}\vspace{-0.1cm}
\end{table}

\begin{figure}[!t]
\centering
\includegraphics[width=0.46\textwidth]{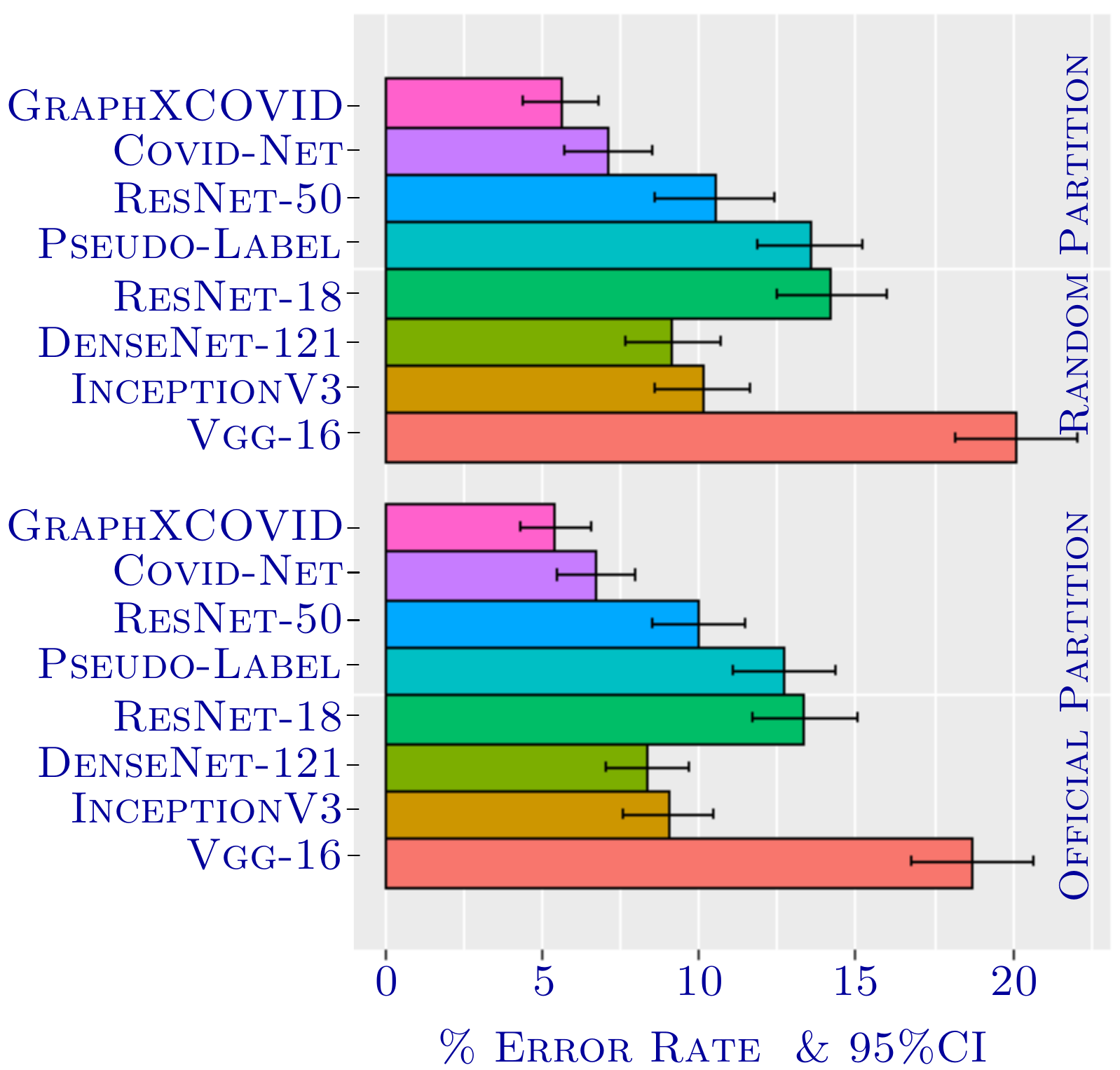}\\ 
\caption{{Error rates on the full dataset. Performance comparison error-wise along with confidence intervals of the compared techniques and ours. Top plot displays results from a random partition, while bottom plot corresponds to the official partition. Our technique reported
the lowest error rate (95\%CI) while using only 30\% of labelled samples.}
}
\label{fig::errorRate2}
\end{figure}

\subsection{Results \& Discussion}
\color{black}
We begin by evaluating the different methods on the official COVIDx partition. As a baseline comparison, we consider six supervised techniques: VGG-16~\cite{Simonyan15},  ResNet-18 and ResNet-50~\cite{he2016deep}, InceptionV3~\cite{szegedy2016rethinking}, DenseNet-121~\cite{densenet2016} and COVID-Net~\cite{covidnet}. To the best of our knowledge, there exists no semi-supervised technique dedicated to COVID-19 identification that we can compare. Hence, for the sake of fairness, we  also adapted one semi-supervised technique, Pseudo-Labelling~\cite{lee2013pseudo},  that has a philosophy close to ours but builds on different principles
including a minimum entropy criterion. The supervised methods use the full training set, whereas the SSL techniques only consider 30\% of the labelled set.

We provide a detailed quantitative analysis to understand the performance of the different techniques. The per class metrics across the official data partition are thus reported in Table \ref{table:comparisonOfficialPartition2}.
Concerning positive predictive values, we observe that our GraphXCOVID approach performs the best for healthy and pneumonia classes and it readily competes with COVID-Net in the COVID-19 class. Due to design limitations, VGG-16 performed the worst, whereas  ResNet-50 presents performances closer to GraphXCOVID thanks to its residual architecture. We also observe that excepting for the COVID-19 class,  InceptionV3 and DenseNet-121 both sightly  outperform ResNet-50.
In order to show the robustness of our technique, we also considered the full COVIDx dataset, a scenario  closer to a real medical setting.  We compared our method with the supervised approach of COVID-Net, the second better method on the official partition. As reported in Table \ref{table:comparisonFullPartition}, GraphXCOVID here performs better for all classes and all  considered metrics, while only using 30\% of the labelled set.

\begin{figure}[!t]
\centering
\includegraphics[width=0.47\textwidth]{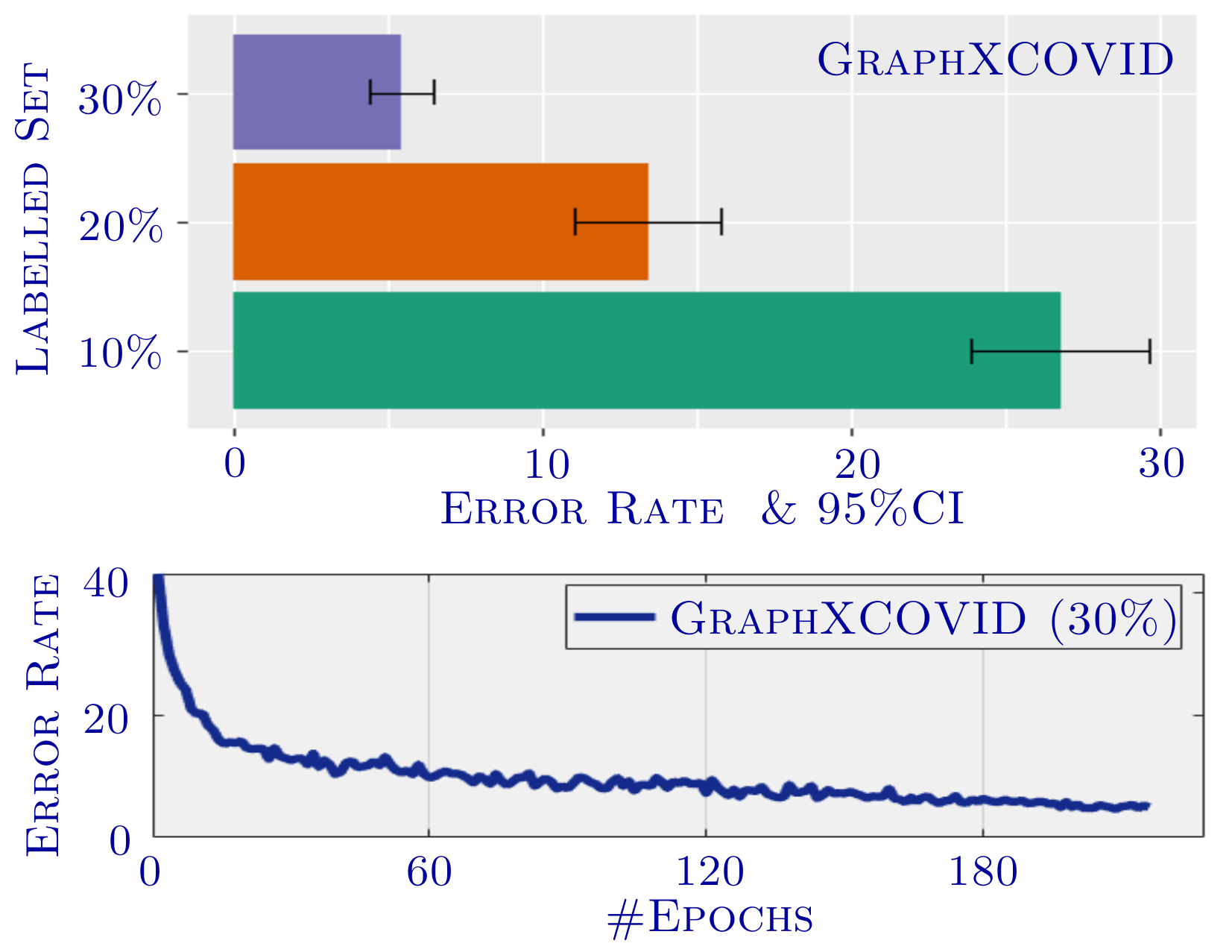}
\caption{From top to bottom. Error rate comparison (95\%CI) of our technique using different label set counts. Error rate vs epochs (i.e iterative process displayed at the center part of Fig. \ref{fig::teasearb})
using 30\% of labels.}\vspace{-0.3cm}
\label{fig::errorRate}
\end{figure}

\begin{figure*}[!t]
\centering
\includegraphics[width=1\textwidth]{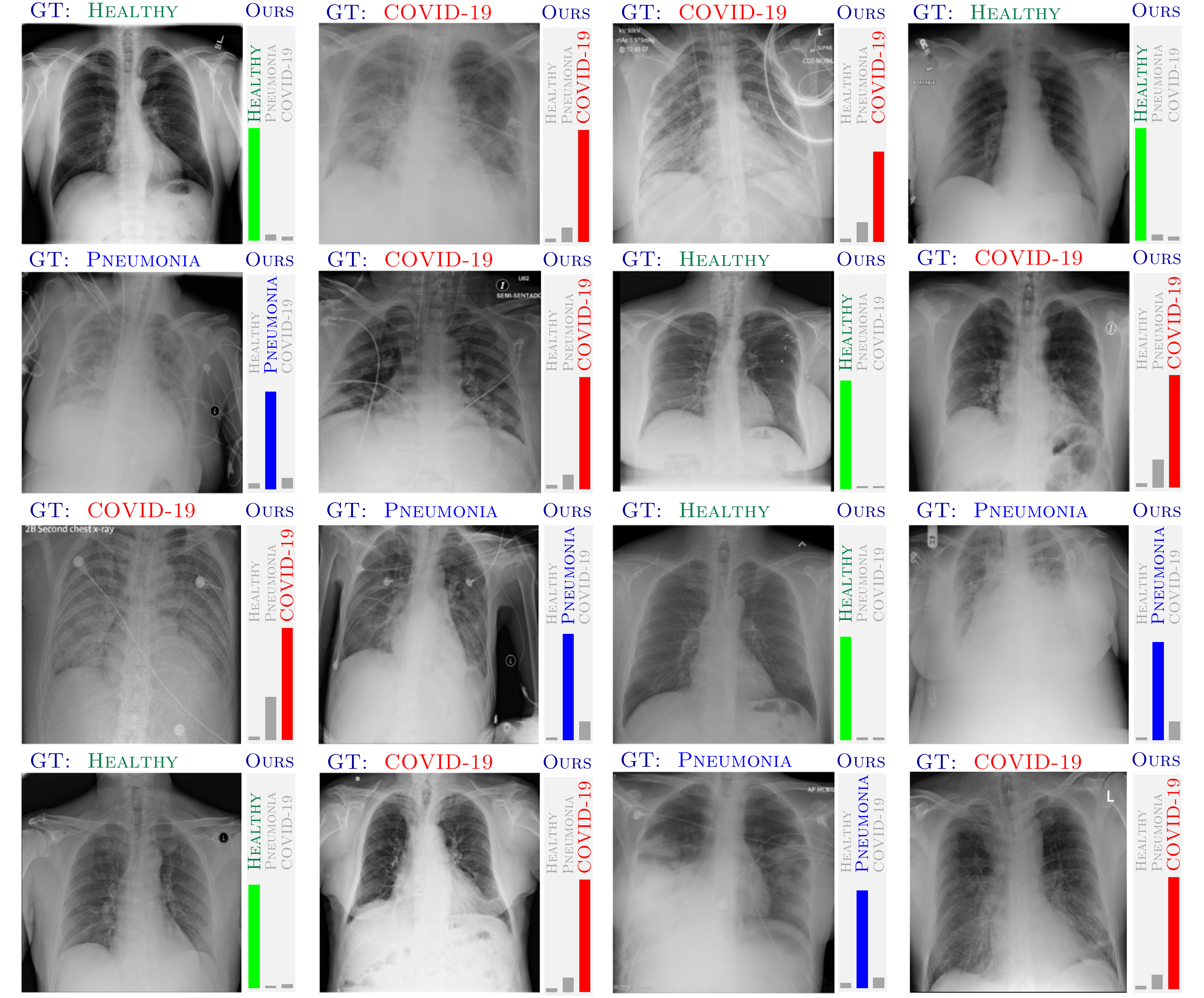}
\caption{Visualisation of results:  the probability score for each class and compared with the human diagnostic (GT).
We see from the confidence measures, that the model has clearly separated normal x-rays from injected x-rays as well as distinguished between pneumonia and COVID-19.} \vspace{-0.3cm}
\label{fig::visu}
\end{figure*}

The second evaluation  is done in terms of sensitivity. As shown in tables \ref{table:comparisonOfficialPartition2} and  \ref{table:comparisonFullPartition}, GraphXCOVID reports the highest values for pneumonia and COVID-19.
For COVID-19, the true positive proportion is significantly higher for  our method (0.94) and COVID-Net (0.91) than for other ones ($\leq$0.88). This observation is confirmed with the  sensitivity results on the highly imbalanced full dataset (see Table \ref{table:comparisonFullPartition}).

To give a view of the relative performance of all techniques, the F1-scores are  respectively reported in Tables  \ref{table:comparisonOfficialPartition2} and  \ref{table:comparisonFullPartition} for the official and full partitions.  It  still underlines that VGG-16, ResNet-50, DenseNet-121 and InceptionV3 are not sufficiently competitive, whereas COVID-Net and our  GraphXCOVID technique are readily performing in a similar level. However, looking deeper into the performance of these two last techniques, we  observe that within a highly imbalanced scenario (see Table \ref{table:comparisonFullPartition}), our approach  outperforms COVID-Net.

We also compare our technique to a SSL technique, with similar philosophy than ours but different in design, Pseudo-Labelling~\cite{lee2013pseudo}. This technique generates the pseudo-labels directly from the network whilst our approach considers labels coming from the diffusion model. From Table \ref{table:comparisonOfficialPartition2}, one can observe that \textsc{GraphXCOVID} offers a substantial improvement over Pseudo-Labelling for all metrics. Overall, we achieve better accuracy with an improvement of ~8\%,  and reduce  the error rate ($\pm1.20$CI) by more than half as displayed in Fig \ref{fig::errorRate2}. The improvement comes from two parts.  The first major benefit is related to pseudo-labels generation. The work of \cite{lee2013pseudo} provides naive pseudo-labels with the network itself, while our technique generates more certain ones that are iteratively updated using both our diffusion models and the network, along with a uncertainty weight. Secondly, our technique also accounts for imbalanced class distribution.
\color{black}

\begin{figure*}[!t]
\centering
\includegraphics[width=1\textwidth]{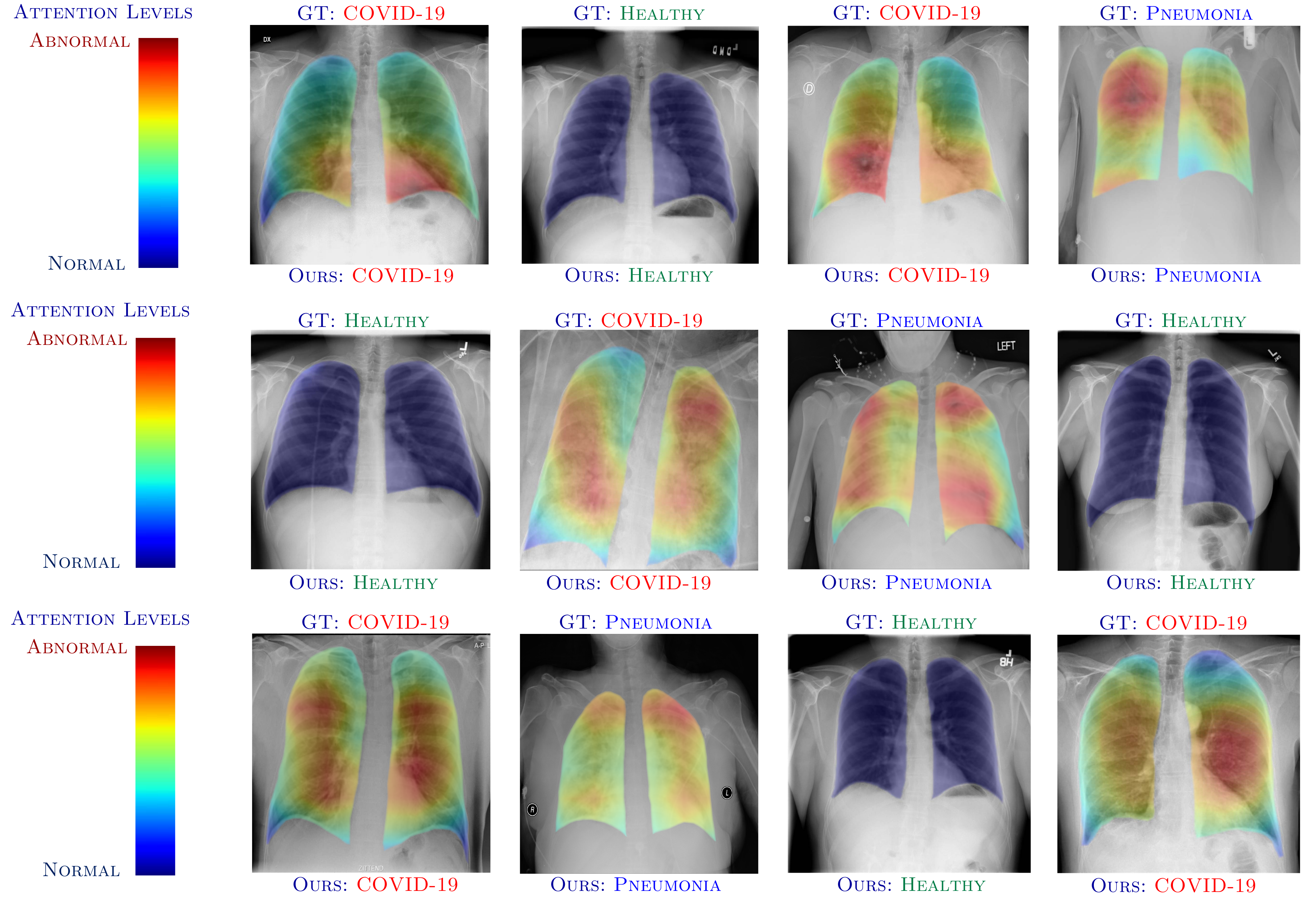}
\caption{Visualisation of the attention maps overlaid on the corresponding chest x-ray image. Our prediction output is also displayed (see bottom part of each output sample) along with the ground truth (GT) which denotes the human consensus prediction (see top part of each output sample). The attention maps highlight abnormal and normal regions to assist the radiologist in making decisions. \vspace{-0.3cm}}
\label{fig::visu2}
\end{figure*}

\color{black}
To further support the previous results and give a global performance view, we compute the error rate and the confidence intervals  for each model.  The results are reported in  Fig. \ref{fig::errorRate2} for the official and a random partition of this set.  For both experiments,  VGG-16  reported the worst performance followed by ResNet-18. Our model performed the best among all the compared models, reporting an error of $5.4\pm1.1$ at the 95\% confidence level. As for the other criteria, COVID-Net ranked second,  reporting an error of the model of $6.7\pm1.23$  at the 95\% confidence level. One can also observe from the top of Fig. \ref{fig::errorRate2} that supervised techniques are highly variable at the change of partition. Such approaches are indeed heavily reliant on the training set being well-representative and balance. In comparison, the  variation is negligible  with our SSL approach.

In order to analyse the robustness  of our model, we run an ablation study for different label counts. In Fig. \ref{fig::errorRate} (top), we present the error rates and confidence intervals obtained by GraphXCOVID with  10\%, 20\% and 30\% of the available labels. Why do not increase the percentage of labels? First, we want to use the lowest possible number of labels. Secondly, we seek to keep the advantage of transductive inference. Indeed, as pointed out in early works e.g.~\cite{joachims1999transductive}, the benefit of a transductive model is decreased when a large number of labels is considered. This effect was observed in our experiments.
We finally illustrate in Fig. \ref{fig::errorRate} (bottom), the behaviour of our model along iterations. Around 200 epochs were required to reach a stable error rate when considering 30\% of labels.

\begin{figure*}[!t]
\centering
\includegraphics[width=0.95\textwidth]{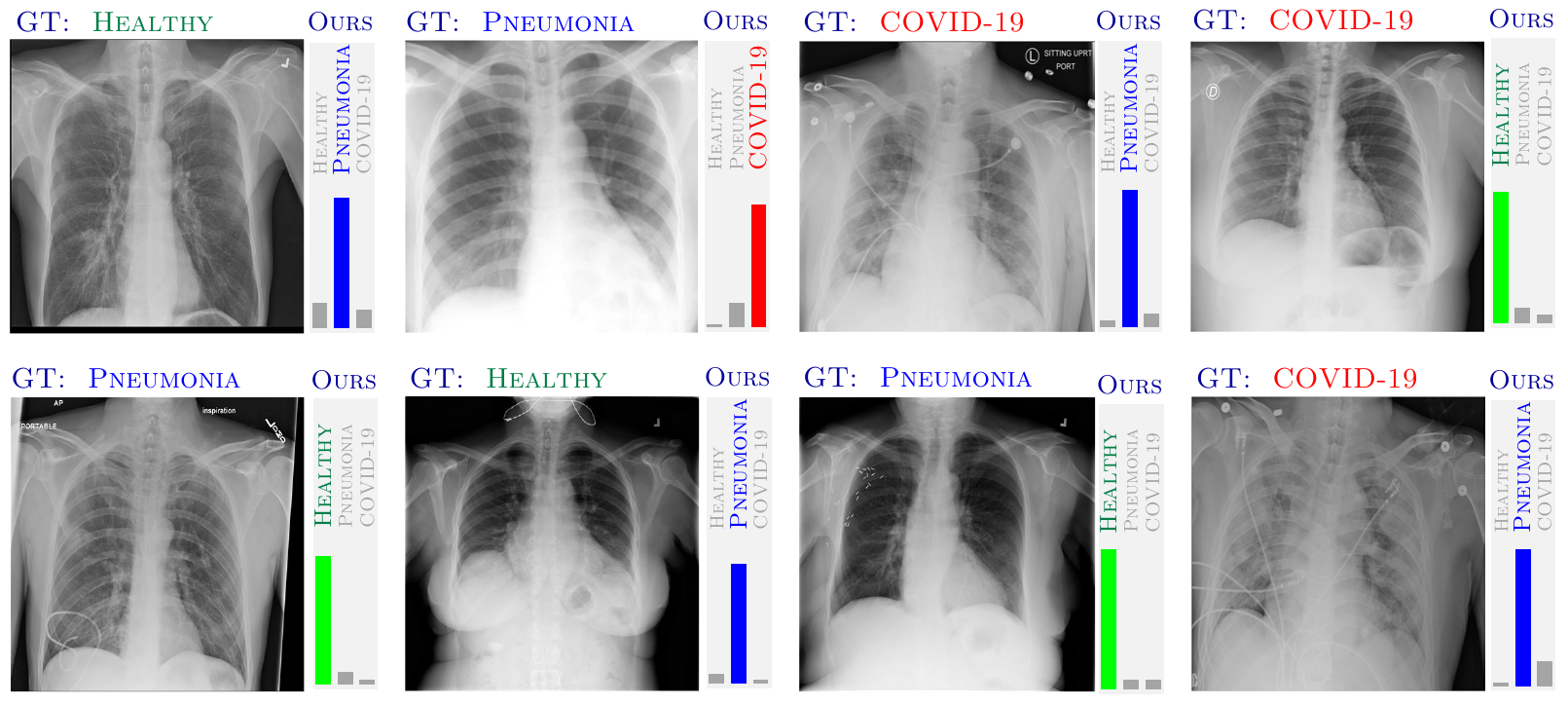}
\caption{{Visual examples of incorrectly classified samples. The results display the predicted probability score and the comparison with the human diagnostic (GT: ground truth).   }} 
\label{fig::failureCases}
\end{figure*}

\begin{figure*}[!t]
\centering
\includegraphics[width=0.85\textwidth]{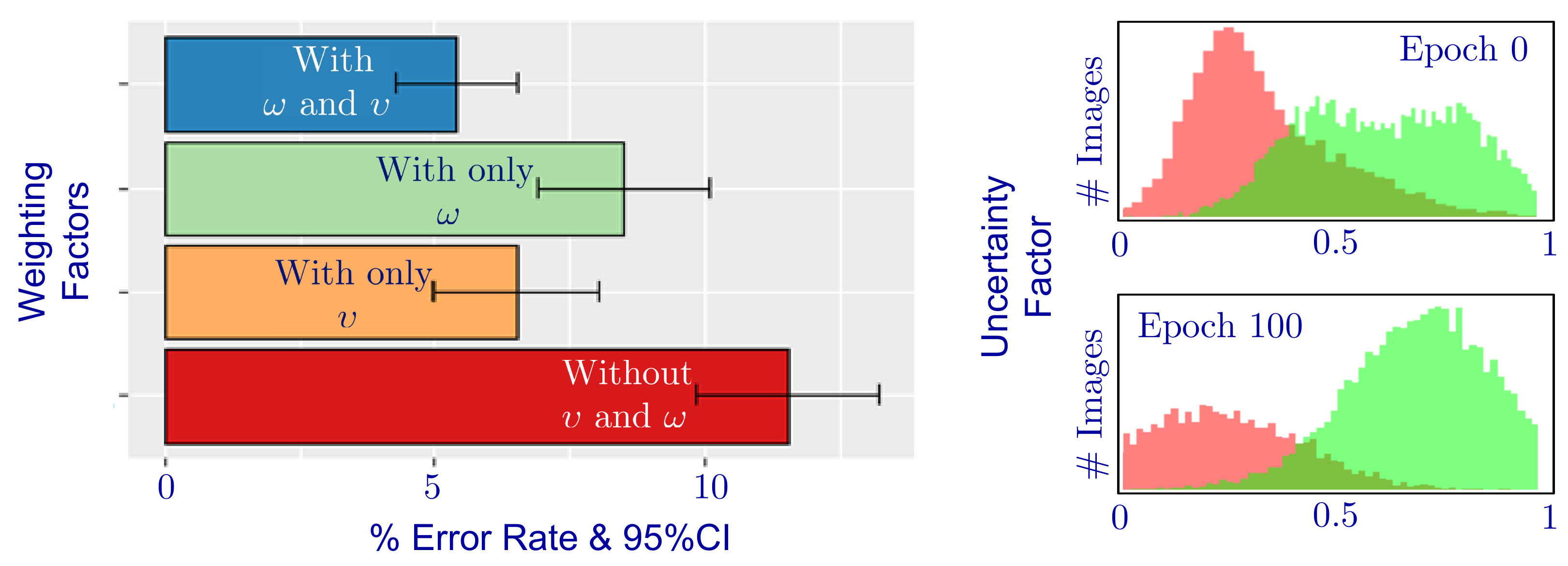}
\caption{{(Left side) Ablation study on the influence of the weighting factors, $\upsilon$ and $\omega$, in the performance of our model. The plot displays the error rate (95\%CI) using 30\% of the labels. (Right side) Effect of the uncertainty mechanism in the pseudo-labels. Correct pseudo-labels are displayed in green and incorrect ones in red.}} 
\label{fig::ablation1}
\end{figure*}

A visual illustration of the result is next provided in Fig. \ref{fig::visu}, where the probability scores of our technique are reported and compared with the human prediction (GT). One can see that the obtained classifier $f_\theta$ easily differentiate between classes.
However, the probability scores (from Fig. \ref{fig::visu}) are not enough to assist the radiologist in making the decision. To accommodate with this issue, we use a Gradient-weighted Class Activation Mapping~\cite{selvaraju2017grad} type solution to highlight abnormal and normal areas in the lungs, in which  Pneumonia and COVID-19 are linked to abnormally regions. Samples outputs of the attention maps are displayed in Fig. \ref{fig::visu2}  and compared with the  human prediction (GT). The attention maps aims to accommodate with the mental model on how the radiologists work in a clinical scenario. Therefore, we project the attention only in the lungs areas. This tool is designed to help, in a friendly user-interface, the radiologist judging whether the diagnostic is correct or not, and in consequence to accelerate the decision.

Additionally in Fig.~\ref{fig::failureCases}, we also display some misclassified samples. The intuition behind these cases is as follows. Firstly, the inherent complex appearance of the pathologies projected in the chest X-ray imposes a challenge in the predictive capability of the model. Secondly, the outcome is also effected by the difference in acquisition protocol and vendor machines, which introduce artifacts, blurry effects and noise in the chest x-ray images. This is translated to a tail distribution problem that affects the model's capability to make predictions. However, we  remark that the generalisation error is reduced with the proposed pseudo-labelling and uncertainty mechanism. In particular, we reported less missing cases of COVID than the compared techniques (see the per class result in Tables~\ref{table:comparisonOfficialPartition2} and  ~\ref{table:comparisonFullPartition}), and our model globally performs better than the compared techniques including when handling external data (see Fig.~\ref{fig::Externaltesting} along with the discussion).

\smallskip
{\textbf{Ablation Study.} Finally and to further support the design of our technique, we performed an ablation study regarding the influence of the weighting factors, $\upsilon$ and $\omega$, in our model.
We first report a performance comparison of our technique, in terms of error rate and using 30\% of labelled data, when considering one, both  and none of the two factors. The results are reported at the left side of Fig.~\ref{fig::ablation1}. We notice that removing both factors (red bar) increases more than twice the error. In contrast, we observe that a substantial decrease in the error rate is achieved when considering both factors (blue bar). In a closer look at the effect of each factor, we observed that while both factors (green and orange bars) indeed improve the performance, the factor with more influence is the uncertainty mechanism ($\upsilon$, see orange bar). We suggest that measuring the uncertainty of the pseudo-labels prevents from obtaining certain incorrect pseudo-labels in early training stages, that are propagated in the next epochs. This effect is illustrated in the right side of Fig.~\ref{fig::ablation1}. In this illustration, we display $\upsilon$ for all the unlabelled samples, and plots reflect the comparison of the pseudo-label prediction with respect of the ground truth. The red and green areas denote the incorrect and correct pseudo-labels (wrt the ground truth). From this figure, we can observe that the certainty in the pseudo-labels is improved as the epochs evolve. This plot support the strength of $\upsilon$ in our model. }

{Additionally, we performed another ablation study to evaluate the performance of our technique with different amounts of labelled data data in the training set. This is illustrated in Fig.~\ref{fig::suppl1}. We observed that  the results using 30\% of the labelled set is a good trade-off between number of labels vs performance due to the transductivity behaviour.  In line with earlier works in transduction e.g.\cite{vapnik1998statistical,joachims1999transductive},  a significant increase in performance is gained from an initial increase in the labelled percentage, from 5\% to 30\%. After $30\%$, the introduction of additional  labelled data provides a tiny amount of extra accuracy. This highlights the success of our approach in its core aim, that is using semi-supervised learning to alleviate the need for large labelled datasets for COVID X-ray classification.

\begin{figure}[!t]
\centering
\includegraphics[width=0.35\textwidth]{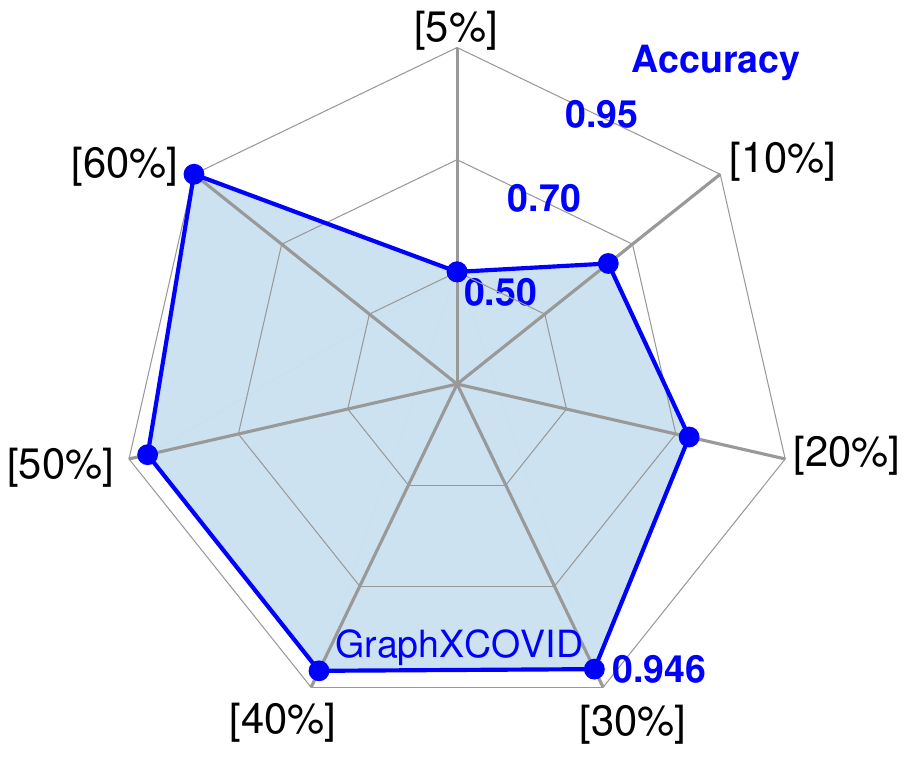}
\caption{{Performance comparison of our technique under different percentage of labelled data. The performance of the model increases with more labeled data, but the increase in performance slows drastically past $30\%$.  }} 
\label{fig::suppl1}
\end{figure}

Moreover, we compared our technique (with and without our weighting factors) against two recent deep SSL approaches: MT  ~\cite{kipf2016semi} and GCN~\cite{tarvainen2017mean}. We considered the same label count of 30\% and used the parameters suggested in these works. The results are reported in Fig.~\ref{fig:comparisonSSL}. We observe that our technique (with the weighing factors) performs the best. MT competes readily with the basic version of our model, i.e.  without uncertainty and balance parameters. Finally, GCN was clearly outperformed by all compared models. The intuition behind our performance gain is three-fold. Firstly, our proposed optimisation model itself enforces the inherent relation between the unlabelled and labelled data. Secondly, the generation of accurate pseudo-labels  with uncertainty estimation increases generalisation. Thirdly, our approach better deals with imbalance datasets where the number of COVID images makes up a tiny fraction of the data.}

\smallskip
\textbf{External Testing. }{ As a final set of experiments, we test the generalisation capability of our technique, and COVIDNet, to \textit{external datasets} by including a performance comparison using the BIMCV-COVID19~\cite{de2020bimcv} dataset (see Subsection 4.1. for details on the experiment). The results are displayed in Fig.~\ref{fig::Externaltesting} and show the error rate at the 95\% confidence level. From the plots, we can observe that COVIDNet exhibits a substantial decrease in performance on the external dataset whilst our technique is more robust in this regard. Particularly, a strong degradation was observed in the COVID-19 class (see left side plot in Fig.~\ref{fig::Externaltesting}). This is an expected behaviour of deep learning models, and is particularly noticed in the medical domain e.g.~\cite{wang2020inconsistent}, as one assumes that the testing set should have similar distribution to the training set.
\textit{Why is our model more robust to external data?} Our technique has been carefully designed to mitigate, at some level, external distributions. This is achieved by discarding irrelevant samples with low confidence scores. More precisely, GraphXCOVID controls the predictive uncertainty on the generated pseudo-labels, which narrows the discrepancy between the external distribution samples. Contrary, COVIDNet is not equipped with any mechanism to accommodate with external distributions.  }

From the aforementioned findings, we emphasise a central message: \textit{the strong performance when using far less labelled data than the compared techniques
is a core strength of our SSL technique. {Moreover, we also underline the robustness of our technique when handling external data.}}
We highlight the value of the vast available unlabelled data in medical domain, and in particular the potentials and benefits for diagnostic COVID-19 disease.

\begin{figure}[!t]
\centering
\includegraphics[width=0.4\textwidth]{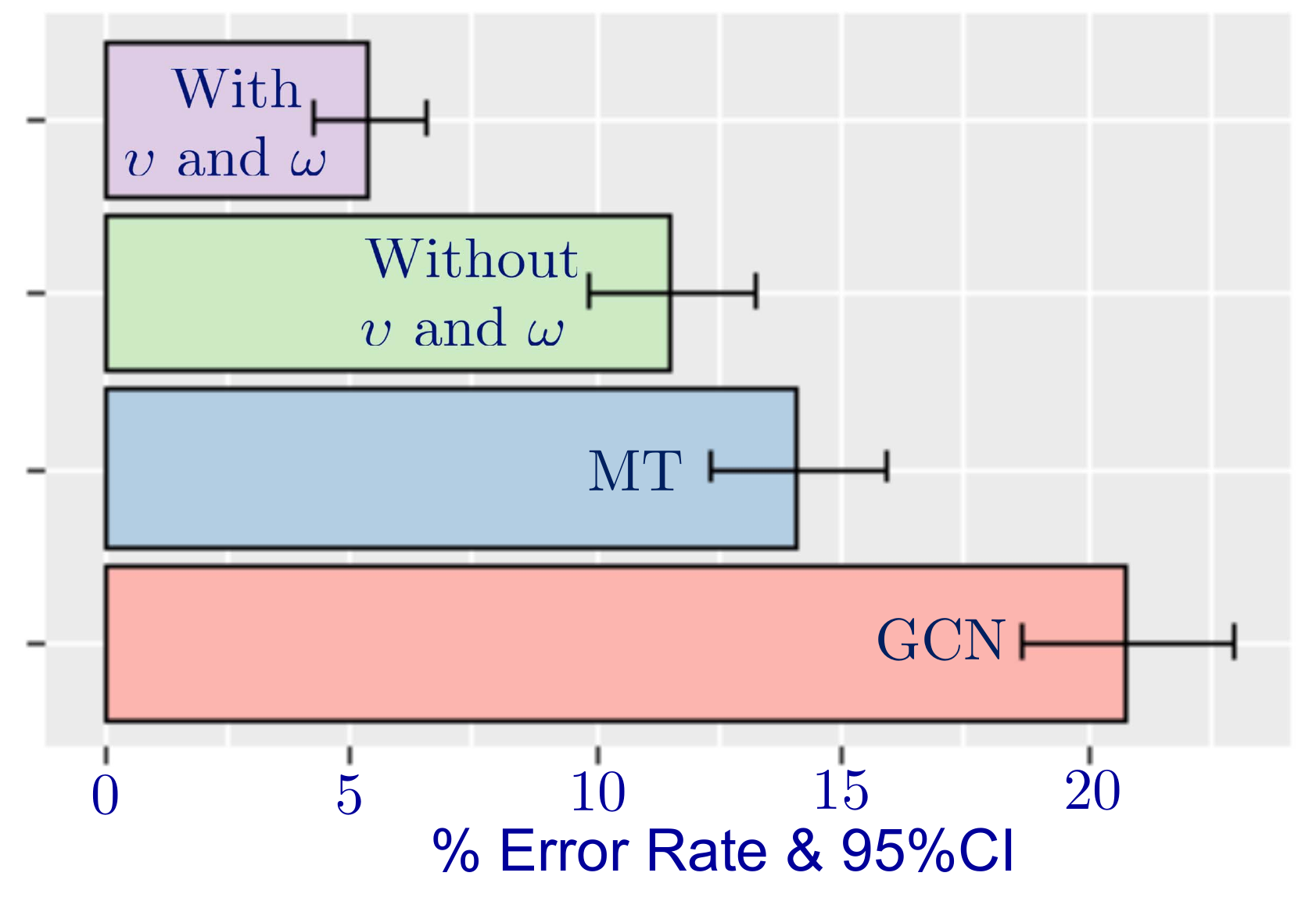}
\caption{{Performance comparison of our technique (with and without including our weighting factors) against two deep SSL techniques GCN~\cite{kipf2016semi} and MT~\cite{tarvainen2017mean}. The results are reported, using 30\% of labelled data, in terms of the error rate (95\%)}} 
\label{fig:comparisonSSL}
\end{figure}
\begin{figure}[!t]
\centering
\includegraphics[width=0.48\textwidth]{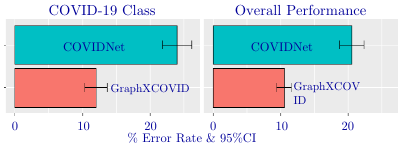}
\caption{{Performance comparison of our technique and COVIDNet using the  BIMCV-COVID19~\cite{de2020bimcv} dataset as a external set. Left side displays the error rate for the COVID-19 class whilst the right side the overall performance of both techniques.  }} \vspace{-0.5cm}
\label{fig::Externaltesting}
\end{figure}

\section{Conclusion}
In this work, we propose a graph-based deep semi-supervised framework for classifying COVID-19 Chext X-ray images based on an optimisation model for label diffusion.  Through the minimisation of a normalised and non-smooth $p=1$ Dirichlet energy, the model generates meaningful pseudo-labels that are iteratively used to update a deep net. To our knowledge, this is the first graph based deep semi-supervised technique for COVID-19 analysis. From our results, we demonstrated that our technique reports higher sensitivity in COVID-19 and better global performances than the current leading deep supervised technique, while requiring a very reduced set of labels. We also provide attention maps as means to visualise the output of our technique. These visualisation aims to assist the radiologist in judging whether the diagnosis is correct. {With this work, we investigate the use of deep semi-supervised learning for novel disease prediction, as for COVID-19. Such approaches alleviate the need for a large labelled dataset, which is costly and time consuming to produce especially for emerging pandemic diseases where both human and monetary resources are stretched thin. {Future work includes the exploration of other strategies to measure the uncertainty, e.g.~\cite{abdar2021review},  and how this can be adapted to the pseudo-labelling setting. Moreover,}
the transfer of our findings in a more thorough  experimental study to test its clinical potential on a larger patient cohort.}

\section*{Acknowledgments}
AIAR gratefully acknowledges support from CMIH and CCIMI, University of Cambridge.
PS is supported by EPSRC and NPL. CBS acknowledges support from the Leverhulme Trust project on
'Breaking the non-convexity barrier', the Philip Leverhulme Prize, the
Royal Society Wolfson Fellowship, the EPSRC grants EP/S026045/1 and
EP/T003553/1, the EPSRC Centre EP/N014588/1, the Wellcome
Innovator Award RG98755, European Union Horizon 2020 research and
innovation programmes under the Marie Skodowska-Curie grant agreement
777826 NoMADS and  691070 CHiPS, the CCIMI and the Alan Turing Institute. AIAR and CBS also thank the team of the project 'AI assisted diagnosis and prognostication in Covid-19' for very helpful discussions.
NP acknowledges H2020 RISE project NoMADS.


\bibliographystyle{IEEEtran}
\bibliography{refs2}

\end{document}